\documentclass[sn-mathphys,Numbered]{sn-jnl}

\usepackage{graphicx}%
\usepackage{multirow}%
\usepackage{amsmath,amssymb,amsfonts}%
\usepackage{amsthm}%
\usepackage{mathrsfs}%
\usepackage[title]{appendix}%
\usepackage{xcolor}%
\usepackage{textcomp}%
\usepackage{manyfoot}%
\usepackage{booktabs}%
\usepackage{algorithm}%
\usepackage{algorithmicx}%
\usepackage{algpseudocode}%
\usepackage{listings}%
\usepackage{graphicx}
\usepackage{float}
\usepackage{subfig}
\usepackage{graphicx}
\usepackage{float}
\usepackage{subfig}

\theoremstyle{thmstyleone}%
\theoremstyle{thmstyletwo}%

\theoremstyle{thmstylethree}%

\begin{document}

\title[Article Title]{FCA-RAC: First Cycle Annotated Repetitive Action Counting}


\author[1]{\fnm{Jiada} \sur{Lu}}\email{lujiada@chinatelecom.cn}
\equalcont{These authors contributed equally to this work.}
\author[1]{\fnm{WeiWei} \sur{Zhou}}\email{zhouweiwei@chinatelecom.cn}
\equalcont{These authors contributed equally to this work.}
\author[1]{\fnm{Xiang} \sur{Qian}}\email{qianxiang@chinatelecom.cn}
\equalcont{These authors contributed equally to this work.}
\author[2]{\fnm{Dongze} \sur{Lian}}\email{liandz@shanghaitech.edu.cn}
\author[2]{\fnm{Yanyu} \sur{Xu}}\email{xuyy2@shanghaitech.edu.cn}

\author[1]{\fnm{Weifeng} \sur{Wang}}\email{wangweifeng@chinatelecom.cn}

\author[1]{\fnm{Lina} \sur{Cao}}\email{caoln@chinatelecom.cn}

\author[2]{\fnm{Shenghua} \sur{Gao}}\email{gaoshh@shanghaitech.edu.cn}


\affil[1]{ \orgname{China Telecom Cloud}, \city{Guangzhou} \country{China}}

\affil[2]{\orgdiv{ShanghaiTech University},   \city{Shanghai}, \country{China}}


\abstract{Repetitive action counting quantifies the frequency of specific actions performed by individuals. 
However, existing action-counting datasets have limited action diversity, potentially hampering model performance on unseen actions. 
To address this issue, we propose a framework called \textbf{F}irst \textbf{C}ycle \textbf{A}nnotated \textbf{R}epetitive \textbf{A}ction \textbf{C}ounting (FCA-RAC). This framework contains 4 parts: 1) a labeling technique that annotates each training video with the start and end of the first action cycle, along with the total action count. This technique enables the model to capture the correlation between the initial action cycle and subsequent actions; 2) an adaptive sampling strategy that maximizes action information retention by adjusting to the speed of the first annotated action cycle in videos; 3) a Multi-Temporal Granularity Convolution (MTGC) module, that leverages the muli-scale first action as a kernel to convolve across the entire video. This enables the model to capture action variations at different time scales within the video; 4) a strategy called Training Knowledge Augmentation (TKA) that exploits the annotated first action cycle information from the entire dataset. This allows the network to harness shared characteristics across actions effectively, thereby enhancing model performance and generalizability to unseen actions. 
Experimental results demonstrate that our approach achieves superior outcomes on RepCount-A and related datasets, highlighting the efficacy of our framework in improving model performance on seen and unseen actions. Our paper makes significant contributions to the field of action counting by addressing the limitations of existing datasets and proposing novel techniques for improving model generalizability.}




\maketitle

\begin{figure}[h]
    \centering
    \includegraphics[width=1\textwidth]{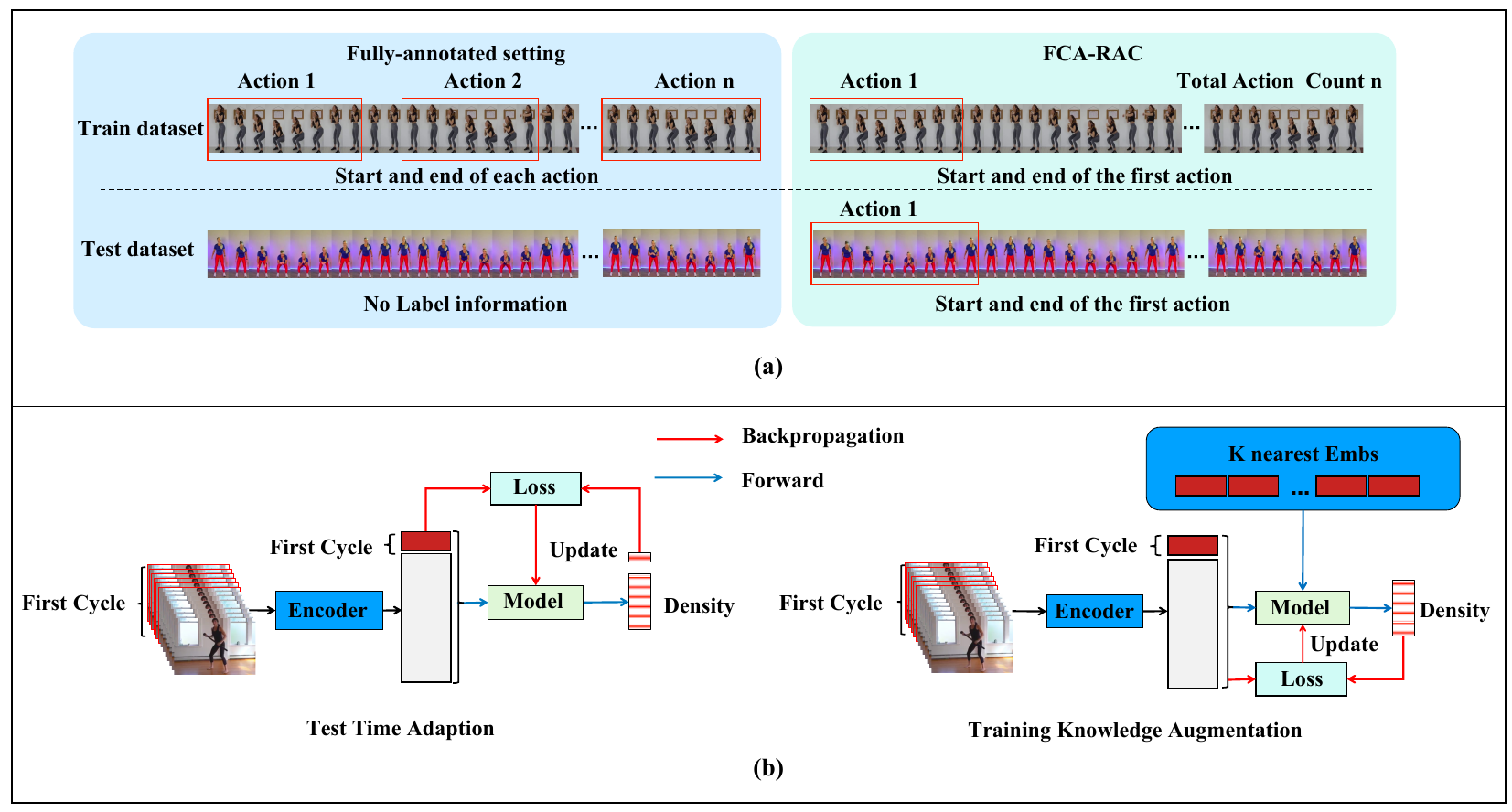}
    \caption{The comparison of the experiment setting. (a) Comparison between  fully-annotated setting \cite{TransRAC} and our FCA-RAC. The red boxes indicate annotated action cycles within the video.  Both methods are trained by annotating the action number  within the video. Besides, in the fully-annotated setting, the start and end frames of each action cycle are annotated as ground truth, whereas in FCA-RAC only  the start and end frames of the first action cycle are annotated. At testing time, while no label information is available  for  the fully annotated setting,  the first action cycle is annotated in our FCA-RAC method. 
    (b) Comparison of model enhancement strategy. In Test Time Adaption \cite{LearningToCountEverything}, the FC-V and V-V model is adapted to each video using the information provided by the  first annotated cycle. In Training Knowledge Augmentation, our FCA-RAC model knowledge is augmented through the first annotated cycle from the training set.}
  \label{fig:setting}
\end{figure}
\section{Introduction}

Repetitive Action Counting (RAC) is a technology that utilizes video capture technology to count the number of repetitions of a specific action. It has potential applications in athlete training evaluation, physical condition monitoring\cite{soro2019recognition}, fitness tracking, and other video analysis studies such as pedestrian detection and reconstruction \cite{lima2021generalizable} \cite{ran2007pedestrian}  \cite{8578418} \cite{ribnick20103d}. Moreover, RAC's ability to accurately count repetitions of an exercise renders it a valuable tool in the field of fitness, enabling individuals to track their progress and measure the intensity of their workout, thus making them stay motivated and focus on achieving their fitness goals. 
To calculate the number of repeated actions in a video, computer vision algorithms such as optical flow or motion estimation can be used to extract features frame by frame and detect the motion of objects in the video\cite{tokta2018fast}. Subsequently, a tracking algorithm can be employed to track the motion of the person and count the number of times the same motion is repeated. These mechanisms enable RAC to provide users with precise and reliable quantitative data on their physical performance, transforming the way we evaluate, train, and optimize human movement.


In previous methods, models were trained on datasets with either coarse-grained annotations (as in  \cite{Countingouttime} \cite{zhang2020context} \cite{zhang2021repetitive}) or fine-grained annotations (as in \cite{TransRAC},\cite{Context-aware}). However, each action-counting dataset only covers a limited range of actions.  In real-world scenarios, there will inevitably be  actions that are not presented in the dataset. As a result, it is reasonable to assume that the model's performance will be sub-optimal when applied to unseen actions. To tackle this problem,  we propose a framework called \textbf{F}irst \textbf{C}ycle \textbf{A}nnotated \textbf{R}epetitive \textbf{A}ction \textbf{C}ounting (FCA-RAC), which contains 4 parts.

The first part is a labeling technique where each training video is annotated with the start and end of the first action cycle and the total count of repetitive actions.  The first action cycle of the test set is also annotated to improve performance on unseen actions. In comparison, previous methods \cite{TransRAC} fully annotate the start and end frames of each action cycle in the training set, without annotation information in the test set. Figure.\ref{fig:setting}a shows the comparison of two methods. 
This labeling technique allows the model to learn the relationship between the first action cycle and subsequent actions, enhancing its ability to predict unseen actions. 
We assume that the model trained on the dataset with our labeling technique  is trying to learn how to match the pattern of the first cycle with subsequent actions in the video.  In other words, our approach seeks to leverage the labeled first action cycle to help the model recognize and predict subsequent actions. In cases where the video contains an action that has not been seen before, the model can still be effective if the first action cycle is annotated. Consequently, this labeling technique significantly enhances the ability of action counting. As long as the training is conducted on a dataset containing several actions, the model can predict any actions, even those that do not appear in the dataset, at the cost of marking the start and end time of the first action. Such an approach has the potential to greatly improve the real-world applicability of action-counting models.

The second part  is a novel video sampling technique. Prior methods  \cite{Context-aware} \cite{TransRAC} relied on fixed frame rates for prediction, where 64 frames are captured from each video in  \cite{TransRAC} for further analysis.  However, this approach may not be sufficient when dealing with videos containing an extreme number of actions. Accurate prediction becomes increasingly difficult as the number of actions exceeds the number of frames, while videos with fewer actions may waste computational resources.  
Another shortcoming  of fixed frame extraction is that some actions are very fast (e.g. skipping rope) while others are very slow (e.g. playing tennis). Fixed frame extraction would lead to the  degradation of model performance due to variations in action speeds. 
To address these issues, we propose a dynamic sampling technique that calculates the sampling rate based on the beginning and end of the first action cycle in each video. The video is sampled according to the speed of the first action cycle. This  approach enables our model to accurately predict the number of actions in different types of videos, regardless of their length or the frequency of the actions they contain.

Variations in the speed of actions can be observed across different activities as well as within a single video. For instance, individuals performing fitness exercises (e.g. front raises) may start with a fast pace, but gradually slow down due to exhaustion.
To tackle this problem, as well as to leverage the benefits of the  sampling technique and annotation methodology, we introduce the Multi-Temporal Granularity Convolution (MTGC) module, the third part of our framework.
The MTGC module employs the encoded feature of  the first action cycle  as the kernel for convolution with the remaining video feature. This enables the model to identify and correspond to the pattern of the initial action cycle with the remaining portions of  the video.  To account for the variations in action speed within the video, the first action cycle kernel is interpolated to different scales. The multi-temporal kernels are then used to convolve across the entire video. This approach facilitates the extraction of action variations across different time scales, enabling the model to capture subtle differences in movement.

To enhance the model performance, previous work focused on adapting the network to the testing examples. Viresh et al. \cite{LearningToCountEverything} propose a Test Time Adaptation (TTA) strategy, that utilizes a few gradient descent updates to adapt the network with the provided testing example.  The procedure of Test Time Adaptation on RAC is illustrated on the left of Figure.\ref{fig:setting}b. However, when it comes to action counting in the video, the extracted feature may vary due to  the pose or angle variation of the person. Relying solely on the first action cycle of a video to match the entire video can lead to a decline in model performance. Hence, utilizing  the similarities of the first action from other videos for fine-tuning and prediction holds the potential to significantly enhance model performance. Additionally, when dealing with an unseen action, the prediction result can also be improved by utilizing the similarities from other videos. 

To this end, we introduced the final part of our framework, called Training Knowledge Augmentation (TKA), as shown in the right of Figure.\ref{fig:setting}b. This strategy utilizes similarities between the first action cycles of videos in the training set for fine-tuning and prediction.  After pre-training, the model constructs an embedding vector space to maintain the first annotated training cycle of each video in the training set. During both fine-tuning and inference stages, the top-k nearest embedding vectors are selected based on their similarity to the input video's first cycle. These vectors are incorporated into convolution kernels to perform MTGC with the input video, improving the prediction accuracy. This strategy attempts to use similar actions within the dataset to enhance prediction accuracy even when dealing with previously unseen actions.   This obviates the need for Test Time Adaptation and enhances the model's performance.

Our proposed method yields satisfactory results on the RepCount-A, CountixAV, and UCFRep datasets. 
Furthermore, we conducted experiments on the UCFRep and QUVA datasets, with the model pretraining on RepCount-A to evaluate the generalization of our method. These results demonstrated that our method achieves competitive performance when applied to previously seen or unseen data.

Our contributions can be summarized as follows:

\begin{itemize}

\item We introduce a  method for annotating repetitive actions that capture the relationship between the first and subsequent actions within a video. 

\item To handle videos with various speeds, we introduce a  dynamic sampling technique that adjusts the sampling rate according to the speed of the first action cycle, creating a more robust representation of the video across various speeds. 

\item  We present a  network module capable of accommodating various action speeds within a single video, which utilizes the multi-temporal scale of the first action cycle as convolution kernels to capture action variations across different time scales. 

\item We present an adaptation strategy to utilize the dataset's annotated first action cycle information. This strategy aims to enhance prediction accuracy by utilizing similarities between actions, regardless of whether the model is predicting previously observed or new actions.
\end{itemize}

\section{Related Work}

 { \bf Temporal Correlation}
Temporal correlation has been widely utilized in video understanding, encompassing areas such as video action recognization  \cite{I3D, 3DResNet}, action detection  \cite{WatchOnlyOnce}, and action localization \cite{chao2018rethinking}.  Vaswani et al.\cite{Attentionisallyouneed} employed attention mechanisms to draw the correlation within the given sequences, while  Junejo et al.\cite{View-independentactionrecognitionfromtemporalself-similarities} propose an approach that extracts temporal correlations between frames using self-similarities.

{\bf Counting in Computer Vision} 
The counting problem has been widely studied \cite{Lian2021LocatingAC}. 
Arteta et al.  \cite{Countinginthewild} focus on counting crowded objects from images using dot annotation. 
 Ranjan et al.  \cite{LearningToCountEverything} addresses the task of counting all objects in an image and proposes an adaptation strategy for testing with limited labeled examples. 
Levy et al.  \cite{Liverepetitioncounting} proposed a method that can count the number of repetitions in videos online. 
Zhang et al.  \cite{Context-aware} tackle the challenge of unknown action cycle lengths by introducing a context-aware and scale-insensitive framework.
Hu et al.  \cite{TransRAC}  introduce a large-scale dataset with fine-grained action cycle annotations, considering realistic scenarios of interruptions and occlusions.

{\bf Temporal  Convolution Network} 
Zheng et al. \cite{zheng2014time} present a multi-channel convolutional neural network (CNN) architecture for effective analysis of multivariate time series data. Their approach processes individual time series with separate CNNs, extracting relevant features that are then concatenated and fed into a novel CNN architecture for further analysis.

Our proposed Multi-Temporal Granularity Convolution module exploits the strong similarities between the first action cycle and the entire video. This is akin to the work of \cite{lu2019class}, where a matching mechanism was proposed to address the counting problem. 
In our proposed method, we utilize the first action cycle as a convolution kernel within the Multi-Temporal Granularity Convolution module, with the aim of matching the pattern of the first action cycle to the entire video. Through this process, we can generate accurate counting numbers for the actions within the video.


{\bf Training Knowledge Augmentation}
Aaron et al.\cite{van2017neural} developed a VQ-VAE method that constructs a  latent embedding space to facilitate the training of encoder and decoder networks. Inspired by this work, as well as research by Zongwei et al.\cite{zhou2021active}, we introduce our proposed TKA, which constructs embedding vector space using the first annotated training cycle from the dataset. During fine-tuning and inference stage, the model will seek the most informative vectors to enhance the prediction, without requiring test time adaptation.

We compare our approach of labeling, sampling, multi-temporal granularity convolution, and training knowledge augmentation strategies on the RepCount, countixAV, QUVA and UCFRep datasets, and empirically show that it leads to better performance on seen and unseen actions.

\begin{figure}
    \centering
    \includegraphics[width=1\textwidth]{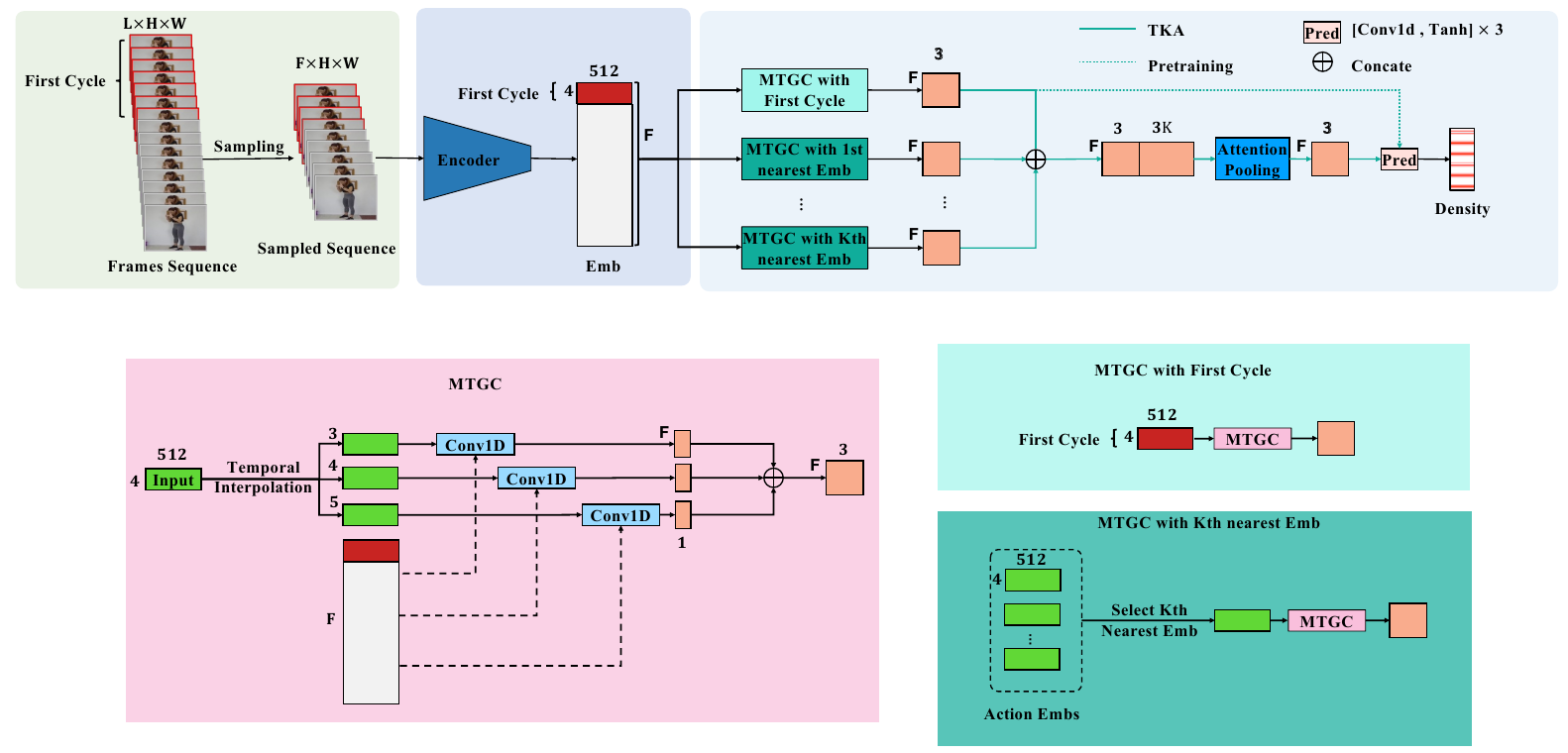}
    \caption{FCA-RAC architecture. The video sequences are sampled according to Sec \ref{subsec:sample}. Then the embedding features are extracted by the encoder.  In the pre-training stage, the first action cycle is scaled to 3,4,5 and used as a kernel to convolve  with the entire videos. After concatenating the feature, we make it pass through the remaining network and output the density map. In the fine-tuning and inference stage, we use the Training Knowledge Augmentation strategy, where the network adds the k nearest  instances of the first action cycle from the training dataset as kernel to convolve with the input video.}
   
    \label{fig:model_sturcure}
\end{figure}

\section{Method}

The aim of this study is to develop an accurate method for counting repetitive actions performed in videos. Figure.\ref{fig:model_sturcure} illustrates the structure of our proposed model.  We start by sampling the video in the pre-training stage. The sampled video is then processed by an encoder and a \textbf{M}ulti-\textbf{T}emporal \textbf{G}ranularity \textbf{C}onvolution module (MTGC), which uses a multi-scaled first action cycle as a kernel to convolve with the entire video.  The resulting feature is passed through a prediction module to obtain the output density map. During the fine-tuning and inference stage, we utilize a \textbf{T}raining \textbf{K}nowledge \textbf{A}ugmentation strategy (TKA) that leverages  similar actions within the dataset to promote the model's performance.

\subsection{Input Sampling}\label{subsec:sample}

As shown in Figure.\ref{fig:model_sturcure}, assuming the input video contains $L$ frames $V=[v_1,v_2,...v_L]$,  the first $N$ frames $V_1=[v_1,v_2,...v_{N}]$ are labeled as the first action cycle.  We sample these N frames to k frames (k = 4 in this article) to represent the action's start, middle, and end stages.  The sampling rate can be calculated as $R=k/N$.   Consequently, the sampled video ( $V_s=[v'_{1},v'_{2},...v'_{F}]$, $F=R*L$) contain ${F}$ frames. 
By sampling all videos according to the speed of the first action, we ensure consistency in action speeds across different videos, thereby mitigating the effect of speed variations. 

\subsection{Encoder}\label{subsec:Encoding}

We extract the video sequence $V_s$ with sliding window step size $\frac{k}{2}=2$, as in Eq.\ref{eq:vk}.

\begin{align}
V'=\{\{v'_1,...v'_4\};\{v'_3,...v'_6\};...\}
\label{eq:vk}
\end{align}

For feature extraction, we employ the Video-Swin-Transformer\cite{Video_SwinT}, which is designed specifically for visual recognition tasks.

Let the video vector $V'$  pass through the Video-Swin-Transformer. For each clip, the size of the extracted features is  $7 \times 7 \times \frac{k}{2} \times 768$, which can then be concatenated to  form a tensor of size $7 \times 7 \times F \times 768$. 

We then apply a conv3D layer with  $ 3 \times 3 \times 512$ filters and ReLU activation, to enhance the combination of spatial and temporal features.
A spatial pooling layer aggregates space information into the channel dimension, resulting in the final temporal context features X as defined below:
\begin{align}
X=[x_1,x_2,...x_F] 
\label{eq:X}
\end{align}


\subsection{Multi-Temporal Granularity Convolution}\label{subsec:MTGC}
To capture the similarities between the first action cycle and the entire video, we introduce the multi-temporal granularity convolution (MTGC) module. The MTGC module leverages the spatio-temporal information contained in the  first action cycle to estimate the count for the entire video.
We use  the first action cycle  $X'=[x_1,x_2,x_3,x_4]$, as a temporal granularity convolution kernel, to convolve with the entire video feature $X$. To account for the variations in action speed within the video, we interpolate the first action cycle kernel to 3 and 5 in different scales, respectively, resulting in three multi-scaled kernels. Each kernel performs temporal granularity convolution on the entire video with stride 1, and their outputs are concatenated to obtain the temporal granularity feature $G\in \mathbb{R}^{F\times 3} $.


\subsection{ Prediction Layer}\label{subsec:predict}
In the pre-training stage, $G$ is passed through the prediction layer, which consists of three 1D convolution layers and the Tanh activation function. The output is the final density map $D=[d_1,d_2,...d_F]$  representing the action period distribution. By summing up the density, we  obtain the action count in the video.

{\bf Density Map}  The advantage of  the density map is the strong interpretability \cite{Liu_2019_CVPR} \cite{9009065} \cite{Tan2019CrowdCV} \cite{10.1007/978-3-030-01234-2_17}, with each frame representing  the action's distribution. To construct the density map of the first action cycle, we follow the procedure in \cite{TransRAC},  using the Gaussian function \cite{guo2011simple}   with a 99\% confidence interval. Since a 99\% confidence interval corresponds to $\mu\pm3\sigma$, we could obtain $G_{\mu,\sigma}(x)$ from $[1,k]$. Then the density value $d_i$ can be calculated as follows:
\begin{equation} 
   d_i = \int_{i+0.5}^{i-0.5} G_{\mu,\sigma}(x) dx, \quad i\in[1,k]
\end{equation}

This approach allows us to construct an accurate density map of the first action cycle to train the model.

\subsection{Training Knowledge Augmentation}
In the proposed Training Knowledge Augmentation (TKA) strategy, we leverage the similarities of the first action cycle between the input video and training set to enhance model performance. After pre-training, we construct an embedding vector space $e\in \mathbb{R}^{T\times k\times D }$ capturing the embedding vectors from the first annotated cycle of each video in the training set, where T is the total number of the training videos, and k is the number of sampled frames of the first cycle (i.e. $k=4$ in this article), and D stands for the dimension of the  embedding vectors.  Each embedding vector $e_i\in\mathbb{R}^{k\times D},i\in1,2,...T$ is obtained by passing the sampled first action of the training video through the Encoder described in Sec.\ref{subsec:Encoding}. During both fine-tuning and inference stages, the top K closest embedding vectors are selected based on their similarity to the temporal context features of the first cycle $X'$ from the input video. The Euclid distance between $X'$ and $e_i$ can be defined as follows: 
\begin{equation}
d_i=||X'-e_i||_2
\end{equation}
Then we sort the distances d in ascending order and select the top $K$ vector with the smallest distance. Therefore, we can express the process of selecting the $K$ closest vector using the formula:
\begin{equation} 
   E=\{e_{(1)},e_{(2)},...e_{(K)}\}  \quad e_{(i)}\in e, d_{(i)}\leq d_{(i+1)}
   \label{eq:Kclosest}
\end{equation}
Here $e_{(i)}$  represents the $i$th closest vector to $X'$  after sorting all values in $d$. Subsequently, each embedding vector $e_{(i)}$ performs MTGC with the entire video, as described in Sec.\ref{subsec:MTGC}. After concatenating the output of the TKA module with the original output of MTGC, we obtain the feature   $G'\in \mathbb{R}^{F\times 3\times(K+1)} $.


To compress the resulting feature, we employ a feature fusion strategy using an attention-pooling layer\cite{Regression}\cite{chen2023self} with learnable weights:

\begin{equation} 
    W=\sigma(Linear(G')) 
\end{equation}
\begin{equation} 
   G''=\sum_{i=1}^{K+1}W_iG'_i
\end{equation}
Finally, the feature of  $G'\in \mathbb{R}^{F\times3} $ is passed through the prediction Layer to generate the density map.


\subsection{Loss}\label{subsec:loss}

To incorporate the label of the first cycle, we crop the first action from the output density map $D \in \mathbb{R}^{F}$, 
which is denoted as $D^{1st}$. We use the Gaussian function $G(x)$ to derive the ground truth density of the first cycle $Gt^{1st}$. 
Therefore, we can calculate the MSE loss (Mean Squared Error) of the first cycle:
\begin{equation} \label{equ:MSEloss} 
   L_{MSE} = (D^{1st}-Gt^{1st})^2 
\end{equation}

Additionally, we compute the MAE loss (Mean Absolute Error)  between the ground truth count $G^{count}$ and the predicted count $D^{count} = \sum(D)$ obtained by summing the density map values:

\begin{equation} \label{equ:MAEloss} 
   L_{MAE} = \frac{|G^{count}-D^{count}|}{G^{count}} 
\end{equation}

The overall loss function is a linear combination of the MSE and MAE losses, weighted by $\alpha$ :

\begin{equation} \label{equ:overallloss} 
   Loss = \alpha L_{MSE}+ L_{MAE}
\end{equation}

\subsection{Inference}
Given a video clip, since its first action cycle is labeled, we first employed our sampling technique to sample the video. Then the sampled video is fed into the model (with TKA strategy), shown in Figure.\ref{fig:model_sturcure}. To apply the TKA strategy, the top K closest embedding vectors from the training set are selected, and each embedding vector  serves as a convolution kernel to perform MTGC with the input video. After concatenating,  the resulting features are passed through the attention-pooling and prediction layers to generate the density map. 
 After applying a linear sum of the density map, we could obtain the predicted value of the action period.

\subsection{Baseline}\label{subsec:baseline}

The setting of our framework is different from the previous one \cite{TransRAC}\cite{Context-aware}. These methods only use the annotation from the training set, and predict the action number in the test set without annotation, while our framework utilizes the annotated first action in the video during prediction. For a fair comparison, we introduce two baseline modules for action counting prediction. Each baseline utilizes temporal context feature $X$ as input and generates a video density map by passing the features through different module types.

\textbf{The first baseline}, named \textbf{F}irst  \textbf{C}ycle \textbf{V}ideo Attention (FC-V), employs an attention mechanism based on the first annotated  cycle. Given the temporal context  features $X$, we use the first action cycle as the query matrix   Q, while the subsequent temporal features are treated as the key matrix  K and value matrix V. These matrices are processed through trained linear layers. 
The cross-attention feature C is calculated by applying the attention equation:
\begin{align}
C=Attention(Q,K,V)=softmax(\frac{QK^T}{\sqrt{d_k}})V 
\end{align}
The cross-attention feature $C$ is then fed into the   prediction module to generate the output density map.

\textbf{The second baseline}, named \textbf{V}ideo to \textbf{V}ideo Attention (V-V), employs a self-attention mechanism to compute the similarity matrix from the temporal context features $X$, similar to the approach used in TransRAC \cite{TransRAC}. V-V is designed to leverage the entire temporal context of the video, which may provide useful information about the frequency and duration of the repetitive actions.   
Specifically, the feature $X$ is multiplied with two weights matrices  to
obtain keys matrix  K and query matrix  Q. Then the similarity matrix $S$ is calculated by $S=f(Q,K)$ where $f$ denotes the dot product. 

Unlike TransRAC\cite{TransRAC}, we apply a $3\times 3$ convolution layer  to the similarity matrix $S$,  yielding the context feature $[32\times F\times F]$.
 Afterward, an adaptive average pooling layer modifies the last axis from $F$ (the frame length) to 16, resulting in the context feature  C. 
Finally, the  context feature $C$ is fed into the   prediction module to generate the output density map. 

\textbf{Test Time Adaptation}. To enhance the performance of both baselines, we incorporate the Test Time Adaptation using the approach from\cite{LearningToCountEverything} . This adaptation strategy adjusts the network to the input video during test time, further improving the accuracy of the estimated count, as shown in the left of Figure.\ref{fig:setting}b. 
Specifically, during test time, the model parameters are frozen except for the   prediction module. We perform several gradient descent steps on the first action cycle of the testing video to optimize the predicted count with the mean-squared error loss defined in Eq.\ref{equ:MSEloss}.

\section{Experiments}

\begin{table*}
\centering
   \begin{center}
      \begin{tabular}{lcccccccc}
         \toprule
         \multicolumn{2}{c}{Annotation}& \multirow{2}{*}{Algorithms} & \multicolumn{2}{c}{RepCount-A}  & \multicolumn{2}{c}{CountixAV}& \multicolumn{2}{c}{UCFRep}\\\cmidrule{1-2} \cmidrule{4-5} \cmidrule{6-7} \cmidrule{8-9}
         Train&Test&& MAE$\downarrow$  & OBO$\uparrow$ & MAE$\downarrow$  & OBO$\uparrow$& MAE$\downarrow$  & OBO$\uparrow$ \\ \hline
         Count&-&Zhang~\cite{zhang2021repetitive} & - & - &  0.331 & 0.43  & \textbf{0.143} & \textbf{0.80}\\ \hline
         \multirow{6}{*}{\begin{tabular}[c]{@{}c@{}}All\\ Cycles\end{tabular}}&\multirow{7}{*}{\begin{tabular}[c]{@{}c@{}}-\end{tabular}}&X3D\cite{X3D} & 0.910 & 0.11  & 0.933 & 0.08 & 0.982 & 0.33 \\
         &&TANet~\cite{TANet} & 0.662 &0.10  & 0.811 & 0.18  & 0.892 & 0.13\\
         &&VST~\cite{Video_SwinT} & 0.575 & 0.13 & 0.792 & 0.25  & 1.122 & 0.03\\
         &&Huang~\cite{I_A_S} & 0.526 & 0.16  & - & -  & 1.035 & 0.02\\
         &&Zhang~\cite{Context-aware} & 0.878 & 0.16 &  - & -  & 0.147 & 0.79\\
         
         &&TransRAC~\cite{TransRAC} & 0.443 & 0.29 &   - & -  & 0.581 & 0.33\\ \hline \hline
         \multirow{3}{*}{\begin{tabular}[c]{@{}c@{}}First\\ Cycle\\+Count\end{tabular}}&\multirow{3}{*}{\begin{tabular}[c]{@{}c@{}}First\\ Cycle\end{tabular}}&FC-V            &    0.344           &       0.36       &    0.392          &   0.56  & 0.305 & 0.56\\
         &&V-V            &  0.322             &       0.36       &     0.379         &    0.55 & 0.211 & 0.69\\  
         &&FCA-RAC & \textbf{0.268}   & \textbf{0.47} &   \textbf{0.330} & \textbf{0.58}  & 0.150 & 0.77\\
         \bottomrule   
      \end{tabular}
   \end{center}
   \caption{ The models above the double line are trained using cycle annotations or only the action counting numbers and evaluated on the label-free test set. In contrast, the two baselines below, as well as our FCA-RAC model, are trained using first-cycle annotations as well as action counting numbers and evaluated on the first-cycle annotated test set.}
   \label{table:Evaluation}
\end{table*}

\begin{table*}
 \centering
      \begin{tabular}{ccccc}
         \toprule
        & RepCount-A & CountixAV & UCFRep & QUVA \\
        \midrule
        Num. of Videos &1041 & 1863 & 526& 100\\
        Duration(s) & 31927&11401 & 3500 &1754 \\
        Duration(s) Min/Max &4.0/88.0 & 1.8/10.0&2.08/33.84 &2.5/64.2 \\
        Num. of Counts.  &15615 &12722 &3506 &1246 \\
        Count Min/Max &1/141 &2/60 &3/54 &1/6 \\
        \bottomrule 
        
 \end{tabular}

     \caption{Dataset statistics of RepCount-A, CountixAV, UCFRep, and QUVA dataset}

   \label{table:dataset}
\end{table*}

\subsection{Datasets and Evaluation Matrices}

We use 4  repetition  counting datasets \cite{Context-aware,TransRAC,zhang2021repetitive,Real-worldrepetitionestimation} to evaluate the effectiveness of our framework. The summary of their statistics is shown in Table.\ref{table:dataset}. 

\textbf{RepCount-A} The RepCount-A dataset \cite{TransRAC} consists of 1041 videos with about 20,000 fine-grained annotations total.  It includes videos of varying lengths and multiple anomaly cases, with each video annotated for the beginning and end of each action period. The average video length in the dataset is 39.35 seconds, significantly surpassing the durations of UCFRep\cite{Context-aware} and CountixAV\cite{zhang2021repetitive}. On average, each video clip in the dataset contains about 16 action cycles. 

\textbf{Countix-AV}\cite{zhang2021repetitive}, a subset of Countix\cite{Countingouttime}, is a dataset sourced from YouTube. It consists of 1863 videos, with 987, 311, and 565 videos allocated for training, validation, and testing, respectively. Since each video only annotates the count of repeated actions, which does not meet our task requirements, we manually annotate the start and end frames of the first action cycle to align with our framework.

\textbf{UCFRep}\cite{Context-aware}, a subset of UCF101, is a dataset that selectively includes 23 action categories out of the original 101, focusing on cyclically performed actions. The dataset contains a total of 526 videos, with 421 videos assigned to the training set and 105 videos designated as the validation set. Each video in UCFRep is annotated with the temporal boundaries of each repetitive action, making it directly applicable to our task.

\textbf{QUVA} \cite{Real-worldrepetitionestimation}, comprises 100 videos exhibiting various activity ranges. Due to its relatively small scale, the QUVA dataset is commonly employed for evaluation purposes. All videos in the QUVA dataset are annotated with the temporal bounds of each interval, enabling us to assess the generalizability of our method.


 To measure the accuracy of our method, we adopted two evaluation metrics following the previous work \cite{Countingouttime}\cite{Context-aware}: Mean Absolute Error (MAE) and Off-by-One accuracy (OBO). MAE represents the normalized absolute error between the predicted count and the ground truth count, while OBO measures the accuracy of the repetition count across the entire dataset. Specifically, for OBO, we considered a video to be counted correctly if the predicted count is within one count of the ground truth; otherwise, it was considered a counting error. The definitions of MAE and OBO are as follows:

\begin{equation} \label{equ:MAE} 
   MAE = \frac{1}{N}\sum_{i=1}^{N}(\frac{|G_i^{count}-D_i^{count}|}{G_i^{count}})
\end{equation}

\begin{equation} \label{equ:OBO} 
   OBO = \frac{1}{N}\sum_{i=1}^{N}[|G_i^{count}-D_i^{count}|\le 1]
\end{equation}
where $G_i^{count}$ and $D_i^{count}$ represent the ground truth and predicted counts of the $i$-th video, respectively, and $N$ is the total number of videos in the evaluation set.

\subsection{Implementation Details}

We implement our method using PyTorch with four NVIDIA V100 GPUs. The Video Swin Transformer tiny  \cite{Video_SwinT} served as the encoder, which was pre-trained on the Kinetics400. Due to the limitation of GPU memory, the parameter of the pre-trained encoder is frozen. 
To train our FCA-RAC model, we normalized the first annotated cycle to $k=4$ frames and resized all frames to $224\times224$. Within a batch, we padded the sampled frames to match the longest selected frames for training purposes.  The  density map in the padded frames  was  set to 0, so the predicted number of actions would be the sum of the "real" density map. 
We used a weighted combination of $L_{MSE}$ and $L_{MAE}$ for the loss function with weight coefficients $\alpha=10$.   We train the model for 16K steps with a learning rate of $8\times 10^{-6}$ and optimized by the Adam optimizer.

After pre-training, we applied TKA to fine-tune the model. Before fine-tuning, we save the encoded feature of the first action cycle of all training videos to construct the embedding vectors. During  each step of the fine-tuning, the top K closest vectors with the first action of the input video are selected to perform MTGC and the subsequent modules to generate the density map. The embedding vectors were updated at every epoch to align them with the output of the model. We perform 1.6K steps during fine-tuning with a learning rate of $2\times 10^{-6}$.

\subsection{Evaluation and Comparison}

To evaluate the performance of our FCA-RAC model, we conducted comparisons with existing methods \cite{X3D, TANet, Video_SwinT, I_A_S, Context-aware, TransRAC} in Table.\ref{table:Evaluation}. The results reported for RepCount-A and UCFRep datasets are taken from \cite{TransRAC, yao2023poserac}. Since the countixAV dataset lacks label information for action cycle boundaries, we manually annotated the start and end frames of the first action cycle.

Our FCA-RAC model demonstrates superior performance on both the RepCount-A and countixAV datasets. In RepCount-A, the model achieves an MAE of \textbf{0.268} and an OBO of \textbf{0.47}, which outperforms the TransRAC\cite{TransRAC} by 0.175/0.18 on MAE/OBO.  Similarly, in countixAV, our model attains an MAE of \textbf{0.330} and an OBO of \textbf{0.58}, slightly surpassing the result in \cite{zhang2021repetitive}. Notably, the FCA-RAC model's performance on the UCFRep dataset is also comparable with the state-of-the-art.

Our model's experimental setting differs from previous methods. While previous methods lack label information in the test set, we utilize annotated first action cycles in testing videos to enhance generalizability. To ensure a fair comparison, we also evaluate the baseline described in Sec.\ref{subsec:baseline}, as shown in Table.\ref{table:Evaluation}. Our model exhibits MAE/OBO performance gains of  0.076/0.11(0.054/0.11), 0.062/0.02(0.049/0.03), 0.155/0.21(0.061/0.08) over FC-V(V-V) on  RepCount-A, CountixAV, and UCFRep dataset, respectively. These results highlight our model's significant superiority over the baseline across all three datasets.
These findings provide strong evidence that our FCA-RAC is an effective model for repetition action counting.

{\bf Generalization}
We evaluated the generalization capabilities of our FCA-RAC model from two perspectives.

First, we conducted experiments on the UCFRep and QUVA datasets using the model pre-trained on the RepCount-A dataset. The results are shown in Table.\ref{table:generalize1}. Our methods outperform the TransRAC\cite{TransRAC} by 0.339/0.23, and 0.512/0.51 in terms of MAE/OBO on UCFRep and QUVA datasets respectively. These results demonstrate the superior generalization capabilities of our method compared to previous approaches when applied to unseen datasets.

Second, we reorganized the RepCount-A dataset according to the guidelines in \cite{TransRAC}, creating new training, validation, and test subsets with disjoint action types. This ensured that the actions in the test set did not appear in the training set. We then evaluated our FCA-RAC model on this newly constructed dataset, and the results are presented in Table.\ref{table:generalize2}. Our method achieves improvements of 0.175/0.18, and 0.210/0.13 in terms of MAE/OBO in the regular setting and resplit setting respectively, which demonstrates that our model also exhibits strong generalization performance on previously unseen actions. These findings demonstrate the strong generalization performance of our model when faced with diverse and unfamiliar action scenarios.

\begin{table*}
 \centering
      \begin{tabular}{ccccc}
         \toprule
\multirow{2}{*}{Method}                                 
                            &         \multicolumn{2}{c}{UCFRep}           & \multicolumn{2}{c}{QUVA}  \\\cmidrule{2-5}
                            &MAE$\downarrow$& OBO$\uparrow$&MAE$\downarrow$  & OBO$\uparrow$\\ \hline
         RepNet\cite{Countingouttime} &0.999& 0.01&- & - \\
         TransRAC\cite{TransRAC} &0.640& 0.32&0.684 & 0.13 \\
         Ours &\textbf{0.301}& \textbf{0.55}&\textbf{0.172} & \textbf{0.64} \\ \hline
        
 \end{tabular}

     \caption{Performance of different methods on UCFRep and QUVA when trained on the same training set of RepCount A.
}
   \label{table:generalize1}
\end{table*}
\begin{table*}
\centering
      \begin{tabular}{ccccc}
         \toprule
\multirow{2}{*}{Method}                                 
                            &         \multicolumn{2}{c}{regular setting}           & \multicolumn{2}{c}{re-split setting}  \\\cmidrule{2-5}
                            &MAE$\downarrow$& OBO$\uparrow$&MAE$\downarrow$  & OBO$\uparrow$\\ \hline
         TransRAC\cite{TransRAC} &0.443& 0.29&0.625 & 0.20 \\
         ours(w/o TKA) & 0.316 & 0.38 & 0.553 & 0.25\\
         ours(with TKA) &\textbf{0.268}& \textbf{0.47}&\textbf{0.415} & \textbf{0.33} \\ \hline
        
 \end{tabular}

     \caption{The experimental result of different methods on two settings of RepCount-A.
}
   \label{table:generalize2}
\end{table*}

\subsection{Ablation Studies}

In this section, we conduct some ablation experiments to evaluate the effectiveness of the designing of the FCA-RAC model. We train the model on the training set of RepCount-A, countixAV, and UCFRep datasets, and then evaluate their performance on the corresponding test set.

\begin{table*}
\centering
      \begin{tabular}{ccccc}
         \toprule
         \multirow{2}{*}{Method} &\multicolumn{2}{c}{RepCount-A}&\multicolumn{2}{c}{CountixAV}\\ \cmidrule{2-3} \cmidrule{4-5}
         & MAE$\downarrow$ &OBO$\uparrow$& MAE$\downarrow$ &OBO$\uparrow$\\\hline
         Baseline(FC-V)& 0.360 & 0.32 & 0.411 & 0.52\\
         TGC (4) & 0.332& 0.37& 0.361 & 0.56\\
         MTGC (3-5) &\textbf{0.316} &\textbf{0.38}& \textbf{0.341} & \textbf{0.57}\\
         MTGC (2-6) &0.355 & 0.37& 0.393 & 0.53\\
         \hline
 \end{tabular}
   \caption{The comparison of FCA-RAC trained with different types of temporal granularity convolution in RepCount-A and countixAV dataset (w/o TKA).  }
   \label{table:Abla2}
\end{table*}

{\bf Multi-Temporal Granularity Convolution.}
We investigated the impact of applying temporal granularity convolution to our model to accommodate speed variations within a video.
Table.\ref{table:Abla2} demonstrates that MTGC with a kernel scale ranging from  3-5 achieves the best results. By using the multi-scale of the first action cycle as convolution kernels, we improve action counting performance by capturing action variations across different time scales within a video.  However, a kernel scale ranging from 2-6 resulted in inferior performance, suggesting that kernels with extreme sizes may not effectively capture the action patterns in the video and disrupt the predicted results. Moreover, compared with the result of the FC-V baseline,  it can be seen that the use of MTGC significantly improves the performance of the action counting task. 

\begin{figure}
    \centering
    \includegraphics[scale=0.4]{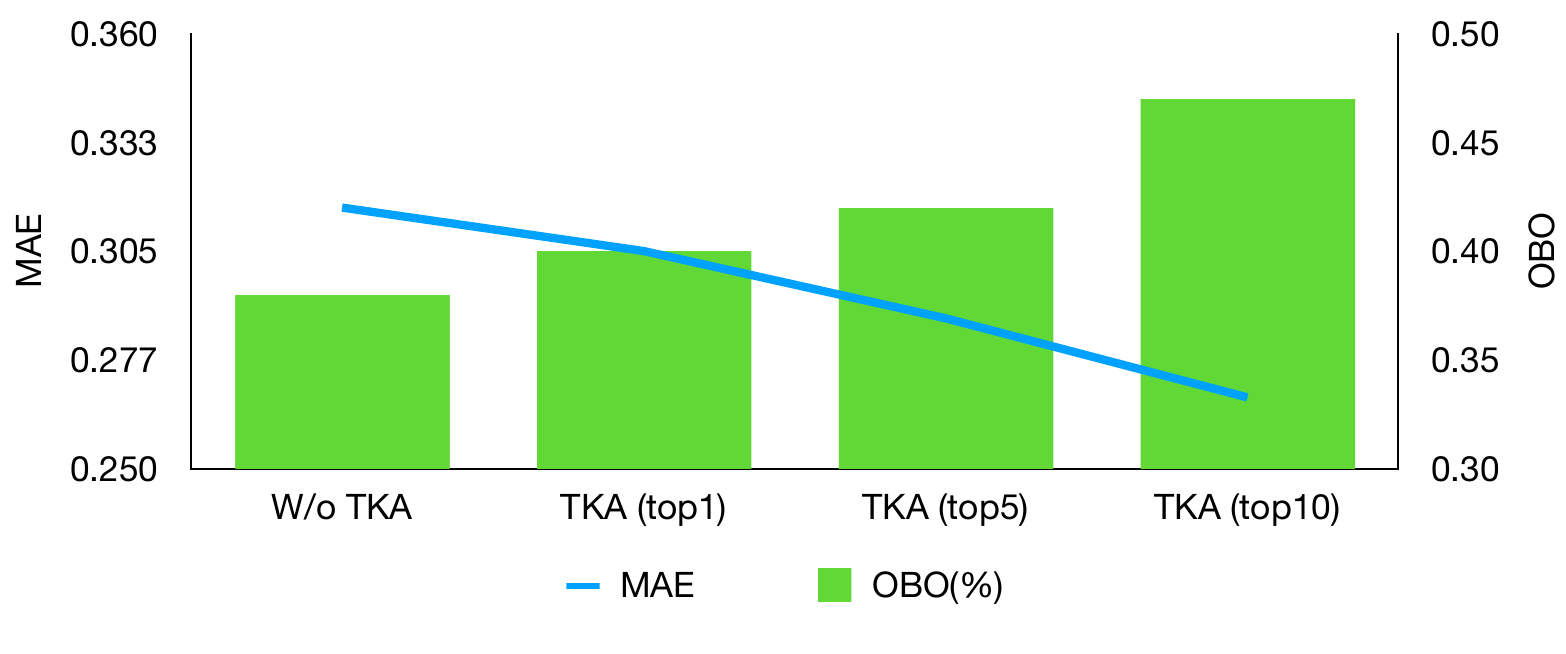}
    \caption{The result of  FCA-RAC with and w/o Training Knowledge Augmentation (TKA) on the RepCount-A dataset. Top1 means using the top1 nearest action cycle from the training set for TKA. }
  
    \label{fig:result_tka}
\end{figure}

{\bf Training Knowledge Augmentation.}
Furthermore, we explored the effectiveness of Training Knowledge Augmentation (TKA). Figure.\ref{fig:result_tka} demonstrates that fine-tuning and inference with TKA surpass the performance without TKA. Moreover, we observed notable performance improvements by increasing the number of action cycles from 1 to 10 during TKA. Table.\ref{table:generalize2}  reveals the performance of  TKA and non-TKA scenarios under different data settings. In the regular setting of the RepCount-A dataset, the TKA strategy gains 0.048/0.09 on MAE/OBO. While in the re-split setting, the corresponding values were 0.108/0.08. 
The results illustrate the strong generalization ability of TKA when dealing with both seen and unseen actions, emphasizing the importance of incorporating multiple cycles in TKA to enhance overall model performance.

{\bf Feature Fusion}
We analyzed the performance variations resulting from different feature fusion strategies on TKA in Table.\ref{table:Abla3}. The max pooling method, which selects the closest feature for TKA, is identical to TKA (top1) shown in Figure.\ref{fig:result_tka}. The result reveals that feature fusion with attention pooling achieves the best performance. This observation indicates that learnable fractions from diverse features within TKA play a role in enhancing the model's performance.

{\bf Loss Term}
As we use two losses to train the model, we compare the performance with different values of $\alpha$. As shown in Table.\ref{table:Abla4}, while training the model exclusively with Mean Absolute Error (MAE) yields effective results, incorporating Mean Square Error (MSE) further enhances performance. Additionally, through experimentation, we determined that the optimal value for $\alpha$ is 10.

\begin{table*}
\centering
      \begin{tabular}{ccccc}
         \toprule
         \multirow{2}{*}{Method} &\multicolumn{2}{c}{RepCount-A}&\multicolumn{2}{c}{UCFRep}\\ \cmidrule{2-3} \cmidrule{4-5}
         & MAE$\downarrow$ &OBO$\uparrow$& MAE$\downarrow$ &OBO$\uparrow$\\\hline
         Average Pooling& 0.285& 0.45& 0.182 & 0.72\\
         Attention Pooling& \textbf{0.268} & \textbf{0.47} & \textbf{0.150} & \textbf{0.77}\\
         Max Pooling & 0.305 & 0.40 & 0.223 & 0.67\\
         
         \hline
 \end{tabular}
   \caption{Performance of different features fusion strategy trained on RepCount-A and UCFRep dataset }
   \label{table:Abla3}
\end{table*}

\begin{table*}
\centering
      \begin{tabular}{cccc}
         \toprule
         Loss & $\alpha$ & MAE$\downarrow$ & OBO$\uparrow$ \\
         \cmidrule{1-4}
         $L_{MAE}$ only & - & 0.327& 0.37\\
         \cmidrule{1-4}
         $\alpha L_{MSE}+ L_{MAE}$ & 1 & 0.293& 0.42\\
         $\alpha L_{MSE}+ L_{MAE}$ & 10 & \textbf{0.268}& \textbf{0.47}\\
         $\alpha L_{MSE}+ L_{MAE}$ & 20 & 0.288& 0.45\\
         
         \hline
 \end{tabular}
   \caption{Comparison of different loss terms of FCA-RAC trained on RepCount-A dataset. }
   \label{table:Abla4}
\end{table*}

\begin{figure*}[!t]
\centering
\subfloat[seen action]{
\begin{minipage}[t]{0.45\linewidth}
\centering
\includegraphics[scale=0.2]{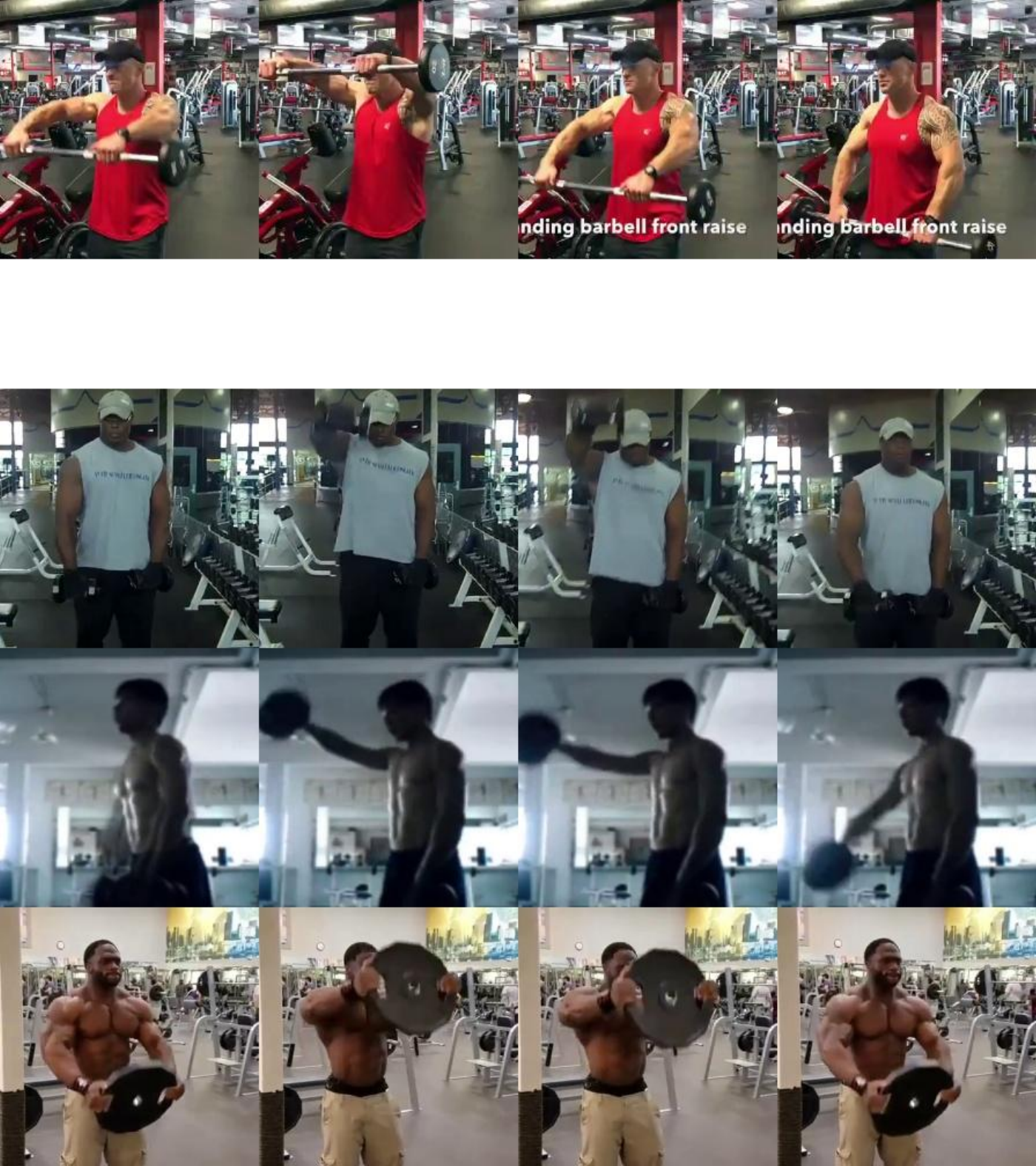}
\end{minipage}%
\begin{minipage}[t]{0.45\linewidth}
\centering
\includegraphics[scale=0.2]{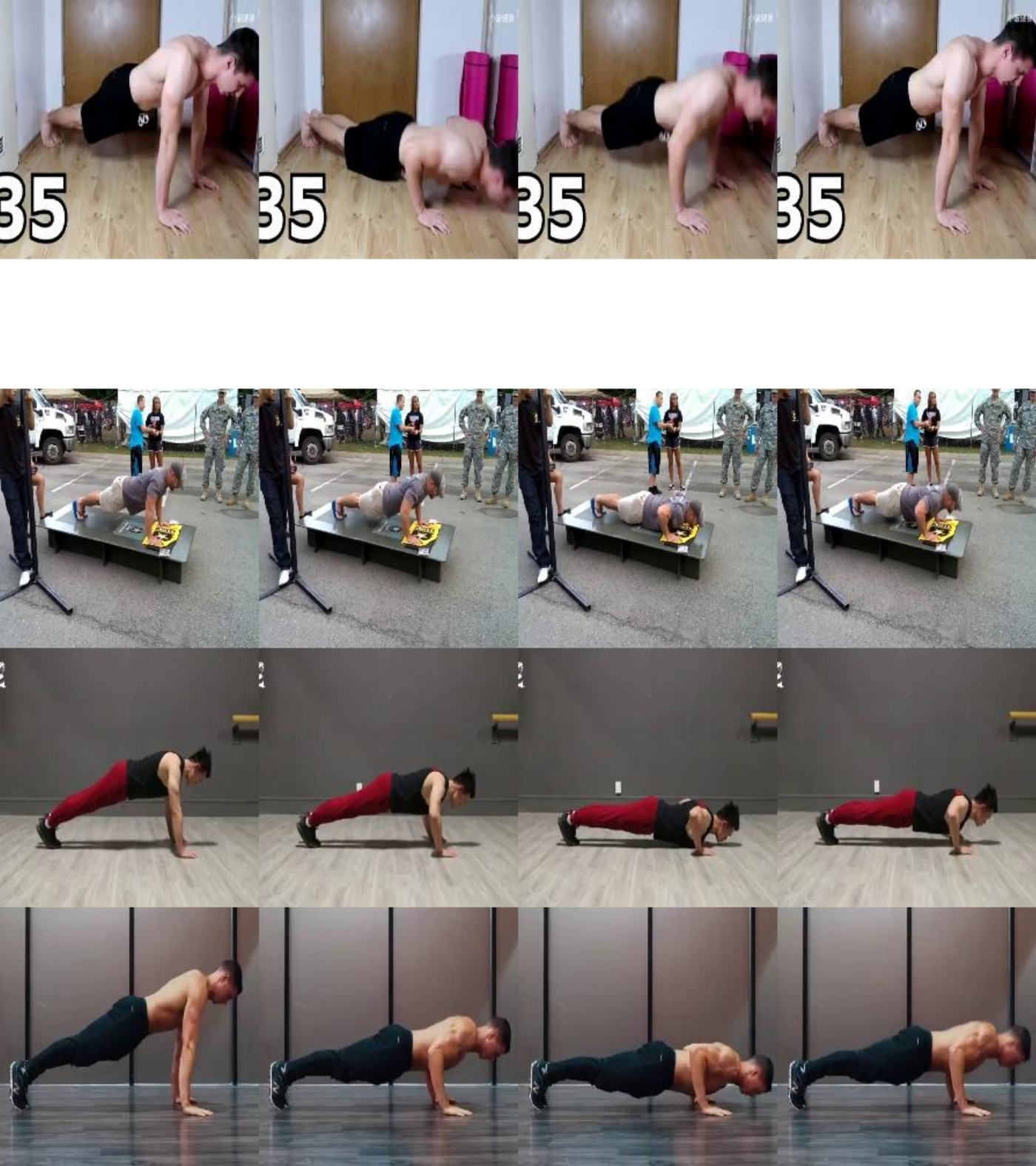}
\end{minipage}%
}\\
\subfloat[unseen action]{
\begin{minipage}[t]{0.45\linewidth}
\centering
\includegraphics[scale=0.2]{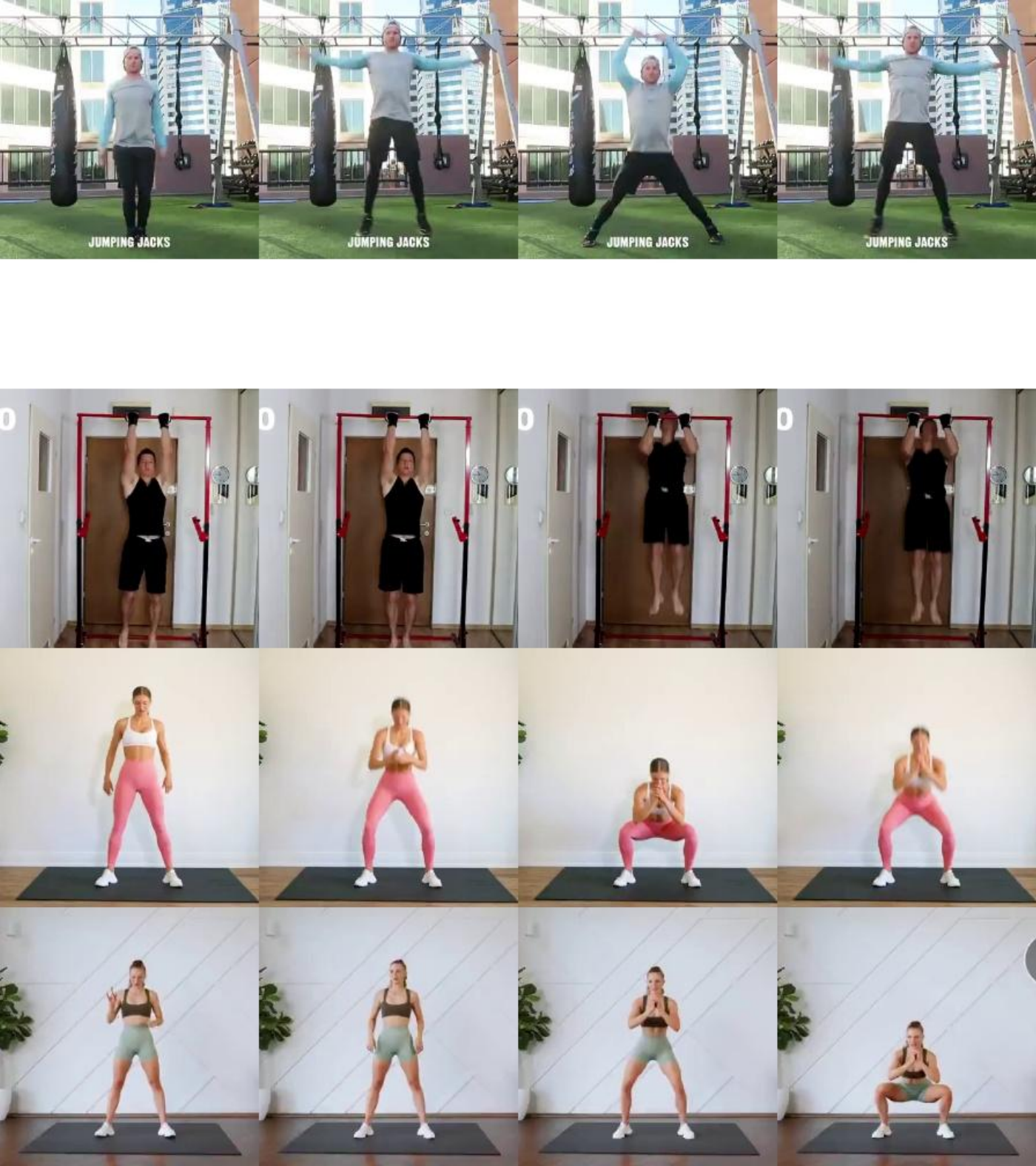}
\end{minipage}%
\begin{minipage}[t]{0.45\linewidth}
\centering
\includegraphics[scale=0.2]{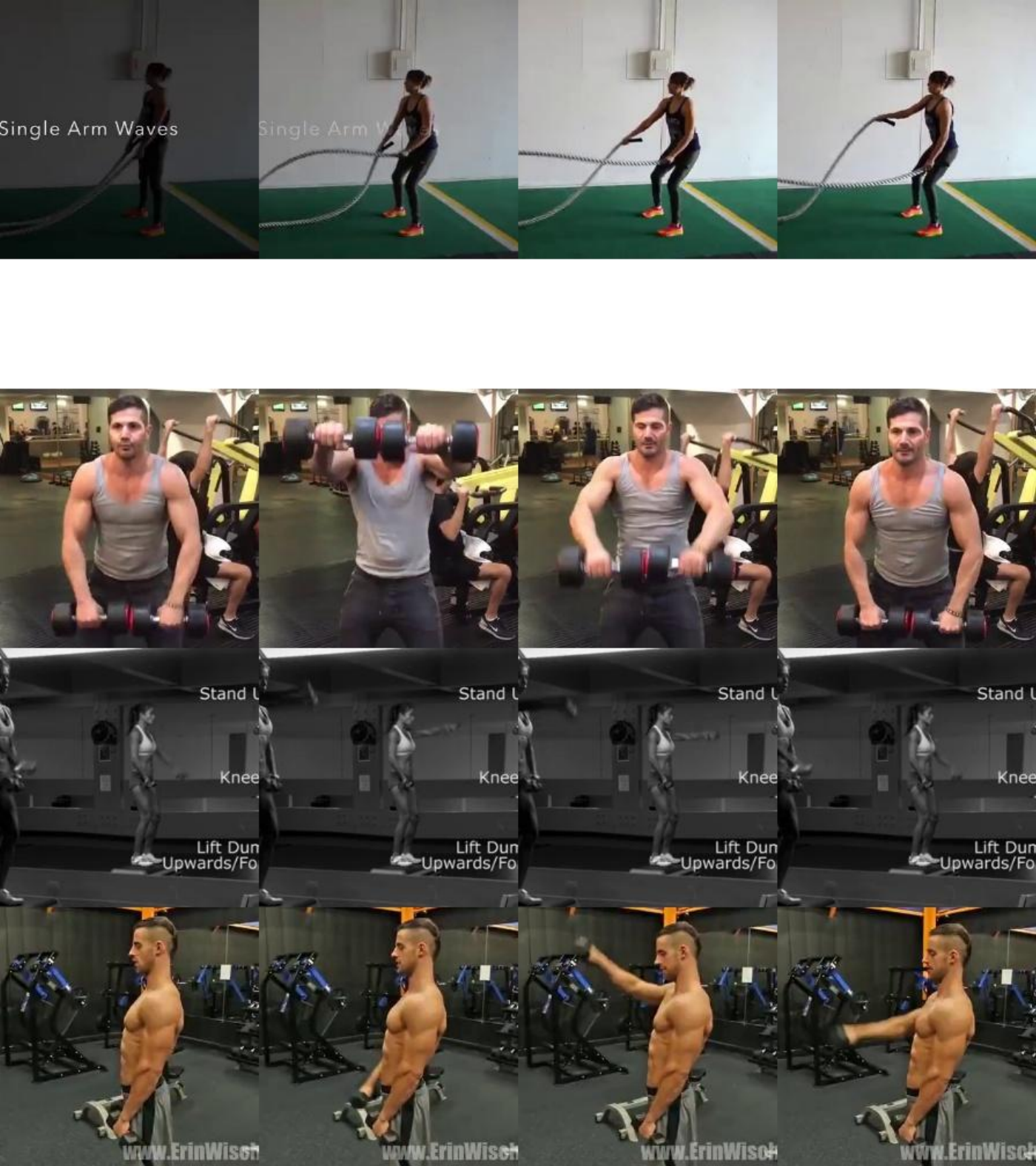}
\end{minipage}%
}%

\centering
\caption{ Visualization of  Training Knowledge Augmentation of seen and unseen action. The top row is the input samples, while the other three rows below are the nearest neighbors selected from the training set by the TKA strategy. For seen actions, the nearest neighbors are coming from the same type of actions. For unseen actions, the nearest neighbors are coming from the samples with similar actions.
}
\label{fig:seen&unseen}
\end{figure*}

\subsection{Qualitative Results}
In this part, we give some qualitative results to better characterize the ability of our framework. 

In Figure.\ref{fig:seen&unseen}, we show some nearest actions selected using our proposed TKA strategy,  given an input action. For seen actions (Figure.\ref{fig:seen&unseen}a), the nearest neighbors belong to the same action type, i.e. front raise and push up. For unseen actions (Figure.\ref{fig:seen&unseen}b),  the nearest neighbors are selected from the samples with similar actions. 
For example, given an input action of jumping jack, the selected actions  are raise up and squat. Similarly, given an input actions rolling scope, the selected action is front raises. This visualization  indicates that the TKA strategy successfully selects the most useful action in the training set to enhance the training and inference performance.

\begin{figure*}[!t]
        \centering
 
        \subfloat[Visualization of good case our model predicted. We can see that the duration of the second pair varies, but the model can still predict the density. ]{
            \includegraphics[scale=0.4]{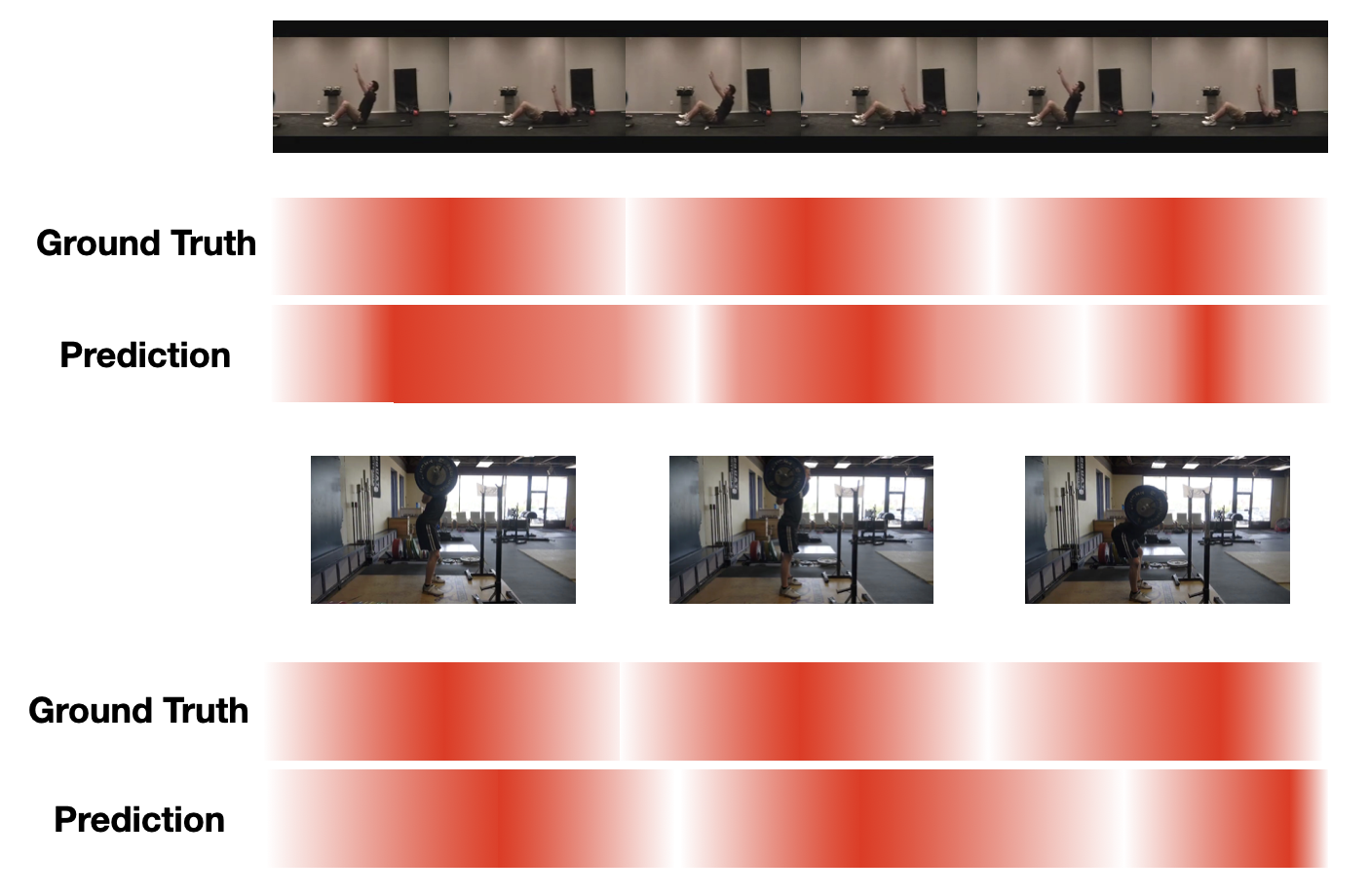}
            }\\
        \subfloat[Visualization of bad case our model predicted. We can see in the second pair that the camera and the people are moving rapidly in the video]{
            \includegraphics[scale=0.4]{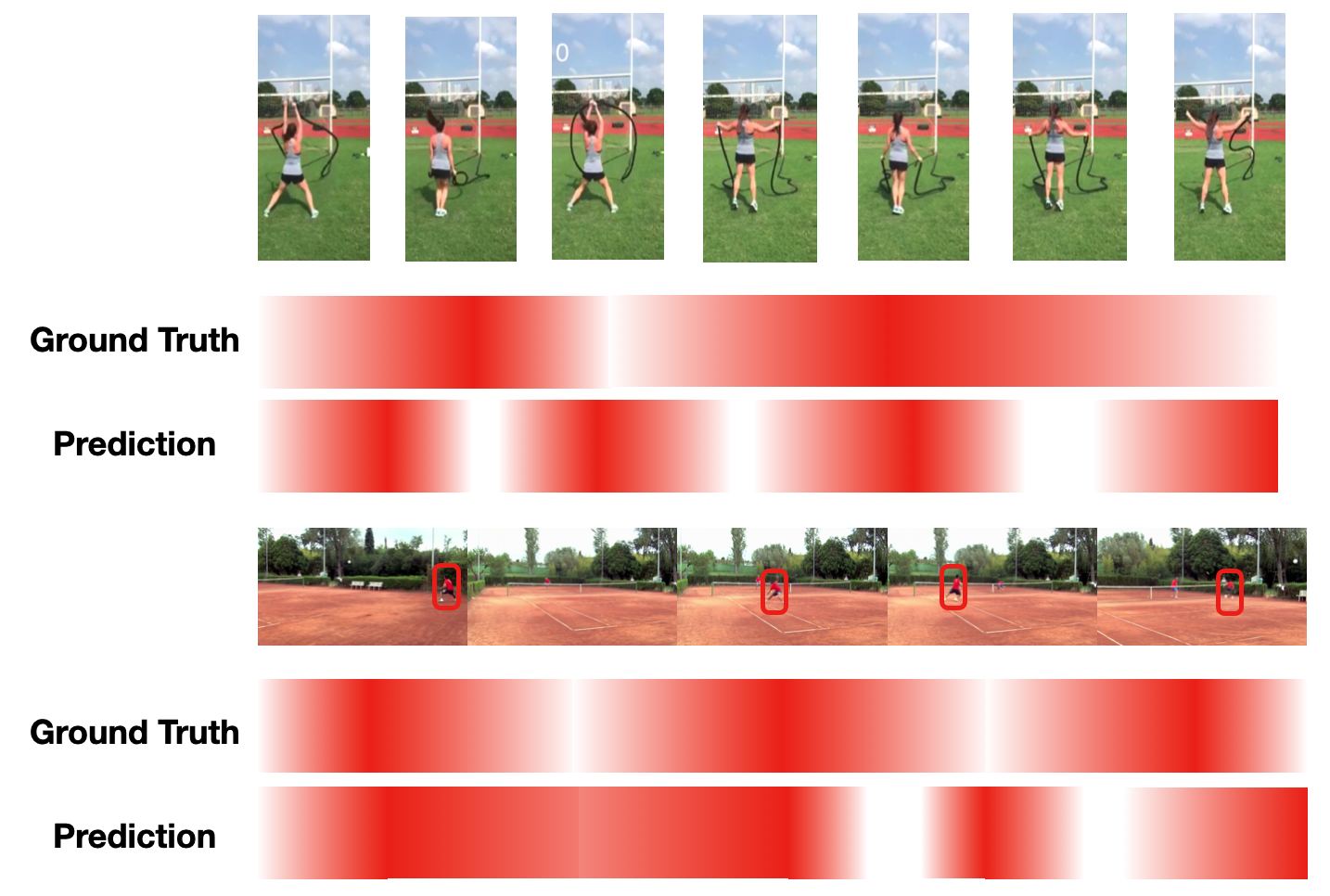}
            }

        \caption{Visualization of the density map of good and bad case our model predicted.}
        \label{fig:good_bad}
\end{figure*}

In Figure.\ref{fig:good_bad}a, we present successful cases achieved by our method.  In the first instance, the subject executes a sit-up with relatively stable frequency, allowing our model to precisely predict the density. In the second example, the individual performs a squat with gradually decreasing speed due to fatigue; nonetheless, our model can also accurately predict the density map.
While our framework performs well on the majority of data, there are still instances of failure, as depicted in Figure.\ref{fig:good_bad}b. In the first pair, a girl performs jumping jacks, and the video speed abruptly slows down in the latter half, approximately one-third of the speed compared to the first half. This big change in video speed disrupts our model's prediction. In the second pair, a boy plays tennis on the ground, and the camera movement is rapid, capturing the boy running and hitting the tennis ball from various angles. These extreme cases can adversely affect our model's performance.

\section{Conclusion}
In  this paper, we introduced a framework called First Cycle Annotated Repetitive Action Counting, which includes a labeling and sampling technique, as well as the MTGC module and TKA strategy. The proposed new labeling technique aims to capture the relationship between the first and subsequent actions within a video. In addition,  our dynamic sampling technique adjusts the sampling rate according to the speed of the first action cycle.
To accommodate various action speeds within a single video, we introduced a Multi-Temporal Granularity Convolution module to capture action variations across different time scales. We also propose Training Knowledge Augmentation, which utilizes annotated first action cycle information within the dataset to improve prediction accuracy, regardless of whether the model is predicting previously observed or new actions.
Our proposed method achieved satisfactory results on the RepCount-A, CountixAV, UCFRep, and QUVA datasets. The extensive experiment shows that our method achieves competitive generalization performance on unseen data.

\newpage
\bibliography{egbib_1}

\end{document}